\documentclass[10pt,twocolumn,letterpaper]{article}

\usepackage[pagenumbers]{iccv} 

\definecolor{iccvblue}{rgb}{0.21,0.49,0.74}
\usepackage[pagebackref,breaklinks,colorlinks,allcolors=iccvblue]{hyperref}
\usepackage{graphicx}
\usepackage{amsmath}
\usepackage{amssymb}
\usepackage{booktabs}
\usepackage{graphicx}
\usepackage{amsmath}
\usepackage{amssymb}
\usepackage{booktabs}

\usepackage{graphicx}
\usepackage{amsmath}
\usepackage{amssymb}
\usepackage{booktabs}
\usepackage{times}
\usepackage{epsfig}
\usepackage{graphicx}
\usepackage{amsmath}
\usepackage{amssymb}
\usepackage{times,enumitem}
\usepackage{epsfig}
\usepackage{graphicx}
\usepackage{amsmath}
\usepackage{amssymb}
\usepackage{mathtools}
\usepackage{tabularx}
\usepackage{makecell}
\usepackage{multirow}

\def\paperID{12642} 
\def\confName{ICCV}
\def\confYear{2025}

\title{MOVi: Training-free Text-conditioned Multi-Object Video Generation}

\author{
Aimon Rahman\textsuperscript{1} \quad Jiang Liu\textsuperscript{2} \quad Ze Wang\textsuperscript{2} \quad Ximeng Sun\textsuperscript{2} \quad Jialian Wu\textsuperscript{2} \\
Xiaodong Yu\textsuperscript{2} \quad Yusheng Su\textsuperscript{2} \quad Vishal M. Patel\textsuperscript{1} \quad Zicheng Liu\textsuperscript{2} \quad Emad Barsoum\textsuperscript{2} \\
\textsuperscript{1}Johns Hopkins University, Baltimore, MD, USA \\
\textsuperscript{2}Advanced Micro Devices
}

\begin{document}


\twocolumn[{
\maketitle
\begin{center}
    \captionsetup{type=figure}
    \includegraphics[width=1\textwidth]{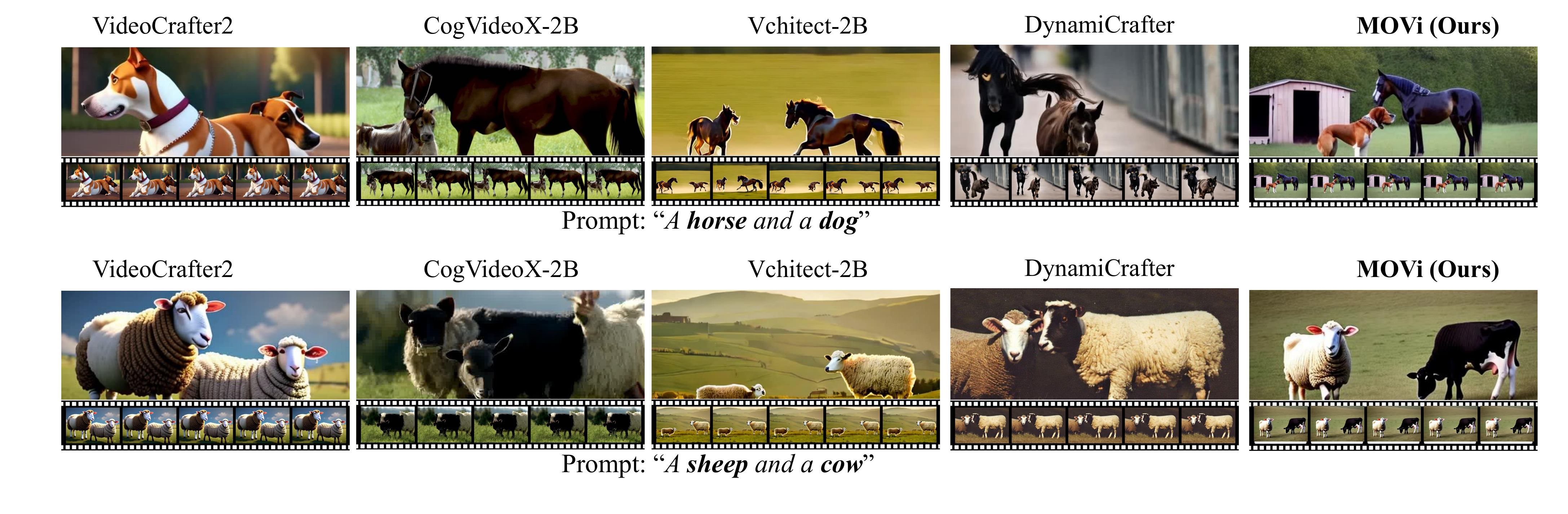}
 \vspace{-20pt}
 \captionof{figure}{Results of multiple-object video generation: Baseline~\cite{chen2024videocrafter2} and other open-source  state-of-the-art models ~\cite{yang2024cogvideox,vchitect,xing2024dynamicrafter} often struggle to generate multiple objects simultaneously. These models frequently prioritize the first object in the prompt or merge multiple objects into a single entity. }
    \label{Fig:first}
\end{center}
}]
\begin{abstract}
Recent advances in diffusion-based text-to-video (T2V) models have demonstrated remarkable progress, but these models still face challenges in generating videos with multiple objects. Most models struggle with accurately capturing complex object interactions, often treating some objects as static background elements and limiting their movement. In addition, they often fail to generate multiple distinct objects as specified in the prompt, resulting in incorrect generations or mixed features across objects. In this paper, we present a novel training-free approach for multi-object video generation that leverages the open world knowledge of diffusion models and large language models (LLMs). 
We use an LLM as the ``director'' of object trajectories, and apply the trajectories through noise re-initialization to achieve precise control of realistic movements. 
We further refine the generation process by manipulating the attention mechanism to better capture object-specific features and motion patterns, and prevent cross-object feature interference. Extensive experiments validate the effectiveness of our training free approach in significantly enhancing the multi-object generation capabilities of existing video diffusion models, resulting in \textbf{42\%} absolute improvement in motion dynamics and object generation accuracy, while also maintaining high fidelity and motion smoothness. \texttt{Code available at: \url{https://github.com/aimansnigdha/MOVi}}


\end{abstract}

\section{Introduction}
\label{sec:intro}

Although video generation models have advanced to produce temporally coherent videos based on various conditions such as text or input images \cite{ho2020denoising,song2020denoising,rombach2022high,villegas2022phenaki,mei2023vidm,ho2022video,vchitect,bar2024lumiere,liang2025movideo,zhang2024show,data_juicer,diff_synth,jtcv}, generating multiple objects and modeling their complex interactions remains a significant challenge. These challenges can be attributed to several factors. First, the foundational text-to-image models that underpin these video generation systems often struggle with intricate prompts involving multiple objects. They lack precise control over spatial relationships, potential occlusions, relative sizes, and similar factors \cite{bar2023multidiffusion,kumari2023multi,liu2023detector}. Second, most video generation models are trained on large-scale web-crawled datasets~\cite{Bain21, xue2022hdvila,chen2024panda}, which, while extensive, are often noisy and may lack sufficient multi-object combinations and interactions. In contrast, stock videos, or pre-recorded, royalty-free video clips sourced from professional video libraries, offer better data quality but typically feature only simple scenes \cite{Bain21}. This limitation makes it challenging to develop generative models capable of handling complex interactions and diverse object relationships within a scene. As illustrated in \cref{Fig:first}, state-of-the-art T2V models often fail to generate two distinct objects. Even when successful, the generated objects occasionally exhibit features overlapping with those of the competing objects, and fail to capture the dynamic motion of object interactions.

Multi-object video generation can be improved by incorporating constraints like bounding boxes \cite{wang2024boximator,li2023trackdiffusion}, which guide models to generate objects within designated regions. However, this approach faces two significant challenges: 1) Additional training is required to learn the new condition, either by fine-tuning a pre-trained model or training from scratch. With new video generation models frequently emerging, it becomes challenging to keep up if each one demands extra training. 2) During inference, it necessitates explicit conditioning, such as supplying bounding boxes, object trajectories, and specific object instances. In contrast, a model conditioned solely on text prompts is much more efficient and easier to use. To address the challenge of multi-object generation without the extensive computational demands of training large models or explicit conditioning in the inference time, we introduce \textbf{MOVi} (\textbf{M}ulti-\textbf{O}bject \textbf{Vi}deo Model)---a novel, training-free framework that enhances the performance of existing T2V models. Our method leverages existing models' capacities in generating realistic single objects and their motions, and extends their functionality for multi-object generation.  
Unlike previous methods~\cite{wang2024boximator,li2023trackdiffusion} that require explicit training with bounding box-conditioned class instances, MOVi only uses text-based conditioning to simplify the process and utilizes large language models (LLMs) as a “director”, guiding object trajectories within generated videos.

It has been shown that in diffusion models, the low-frequency signals in the noise contribute to generating the main objects in a scene, while high-frequency signals primarily represent the background~\cite{wu2023freeinit}. To inject the object trajectory information in a training-free manner, we propose a \textit{{multi-object noise reinitialization}} strategy that manipulates the low-frequency noise components for each object across the frames to control inter-frame content correlation, steering each object's motion and layout across the video sequences. This reinitialization of low-frequency noise components facilitates improved multi-object generation by maintaining spatial consistency. To prevent overlapping/morphing between competing objects, we propose an \textit{{object text token-based attention re-weighting}} method, focusing attention along designated trajectories for each object. This re-weighting strategy reduces the influence of competing objects, refining individual object semantics within their trajectories for enhanced accuracy and coherence in the final video output, as shown in \cref{Fig:first}. We evaluate MOVi against state-of-the-art open-source and commercial T2V models. MOVi significantly achieves a weighted average score of 85\% on dynamic degree and object accuracy in VBench~\cite{huang2024vbench}, outperforming the second-best model, Vchitect-2.0-2B~\cite{vchitect}, which scores 64\%, and substantially improving upon the baseline model's~\cite{chen2024videocrafter2} score of 43\%. In addition, we show that it maintains high video quality and object generation accuracy as the number of object increases. Human preference study demonstrates a clear advantage of our method in generating accurate motion and objects even compared to commercial models.

Our contributions are summarized as follows: 

\noindent 1) We propose a novel, training-free framework called MOVi to improve multi-object video generation without requiring any retraining or additional conditioning as inputs.

\noindent 2) We design the multi-object noise reinitialization and object-text attention re-weighting strategies to control individual object trajectories and enhance the coherence and accuracy of object generation.

\noindent 3) Our extensive experiments demonstrate that MOVi can capture complex object interactions and accurately generate multiple distinct objects as specified in the input prompt.


\section{Related Work}

\noindent\textbf{Video Generation Networks.} Recent years have seen a surge in video generation models, typically built on denoising UNet \cite{blattmann2023align,singer2022make,ho2022imagen,hong2022cogvideo,ho2022video,mei2023vidm,molad2023dreamix,wang2023modelscope}, or diffusion transformer backbones \cite{yang2024cogvideox,liu2024sora}. A common approach is to combine spatial diffusion with temporal layers to capture both spatial and temporal dynamics. However, two key challenges remain: video diffusion models are rarely publicly available, and they demand large-scale training. This rise in video generation extends into the commercial space \cite{lumaAI,runwayml_gen2,runwayml_gen3,minimax,kling}, as these models have the potential to translate a model's understanding of the world into dynamic video content \cite{qin2024worldsimbench,wang2024worlddreamer,cho2024sora}.

\noindent\textbf{Multi-object capabilities of Video Generation Network.} Recent advancements in large-scale video generation networks, trained on vast datasets such as HD-VILLA-100M~\cite{xue2022hdvila}, WebVid-10M \cite{Bain21}, and Panda-70M \cite{chen2024panda}, have shown that these models can implicitly learn multi-object motion from the data itself. By design, they are capable of handling multiple objects, as demonstrated in benchmarks like VBench \cite{huang2024vbench}, which includes a multi-object metric. However, these models still have certain limitations. Approaches like bounding box-guided networks, such as Boximator \cite{wang2024boximator} and TrackDiffusion \cite{li2023trackdiffusion} (based on GLIGEN \cite{li2023gligen}), have demonstrated some success in multi-object generation. Nevertheless, these methods require explicit training and rely on instance and bounding box information during inference. This also limits their open-world generation capabilities. While some attempt to customize multi-object generation by retraining for specific objects \cite{wang2024customvideo,chen2023videodreamer}, this strategy lacks generalization and necessitates retraining for each new set of objects. Similarly, explicit conditioning, such as image or mask conditioning, can improve generation, but it remains dependent on specific conditioning signals \cite{yang2024eva}. In contrast, our approach requires no additional training or object-specific retraining. We leverage the model’s inherent real-world priors of single-object generation and movement to enable generalized and efficient multi-object generation without relying on explicit supervision or object-specific conditions.

\section{Method}
\begin{figure*}[htbp]
\centering
\includegraphics[width=.8\linewidth]
{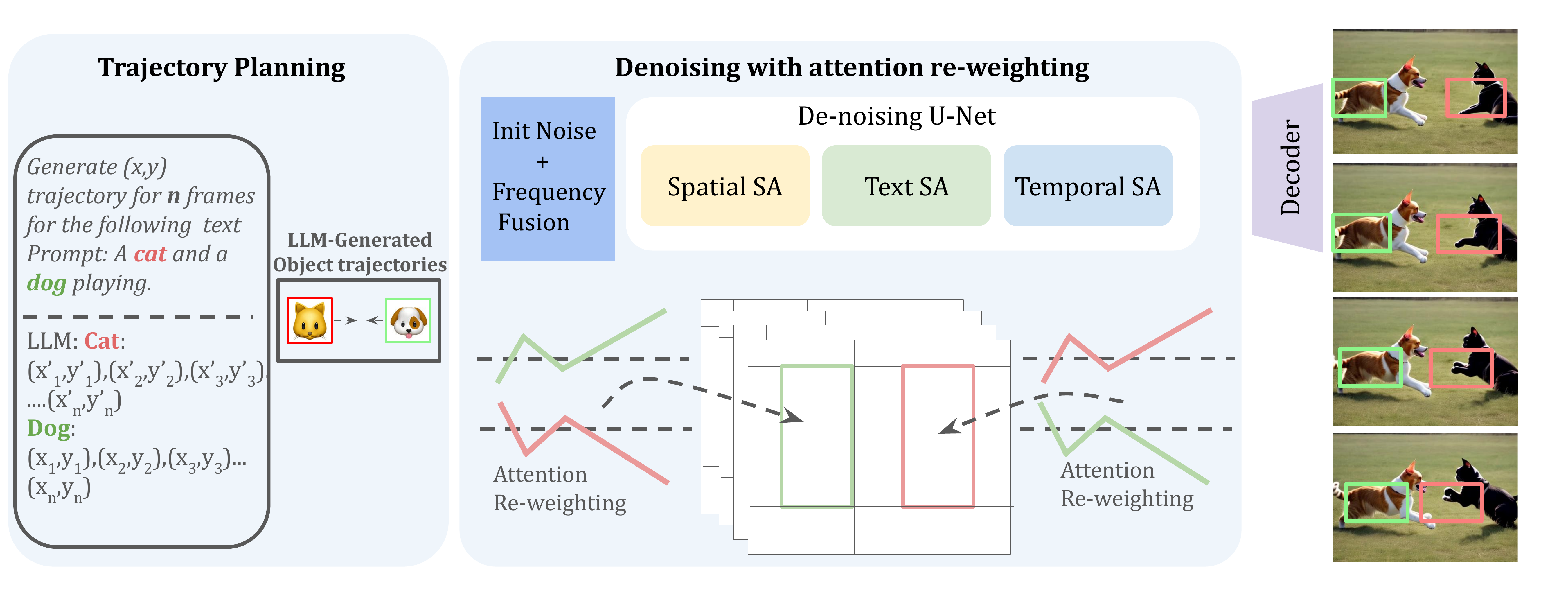}
\vspace{-20pt}
\caption{Pipeline of the proposed \textbf{MOVi} framework. \textbf{SA} stands for the self-attention. In the first stage, an LLM acts as a director, generating object trajectories from the input prompt and specified number of frames. These object trajectories are then used to reinitialize the noise, with masking applied based on the noise's low and high-frequency components. The noise is passed through the network to generate videos with multiple objects. During the iterative denoising process, attention re-weighting is applied to specific bounding boxes based on the prompt to eliminate unwanted object influences and refine the output.
}
\label{Fig:train}
\end{figure*}

Our method exploits the cross-attention mechanism in video diffusion models and the robustness of low-frequency signals to additive noise, which aids in preserving layout and shape from initial noise. Additionally, we harness the LLM’s capability as a trajectory planner, aligning it with spatial and temporal structures. Details are provided in this section. The complete pipeline is illustrated in Fig. \ref{Fig:train}.

\subsection{Preliminaries}
\noindent \textbf{Text-to-Video Generation Models.} In video diffusion models, cross-attention mechanisms are usually employed to integrate text-based conditioning for generating coherent video sequences \cite{wang2023modelscope,chen2024videocrafter2}. Let $\mathbf{X} \in \mathbb{R}^{T \times H \times W \times C}$ denote the spatiotemporal features extracted from the video, where $T$ represents the temporal dimension, $H$ and $W$ the spatial dimensions, and $C$ the feature channels. The text prompt is embedded into a sequence of embeddings $\mathbf{E} \in \mathbb{R}^{L \times d}$, where $L$ is the number of text tokens in the prompt, and $d$ is the embedding dimension. The cross-attention mechanism computes attention weights by aligning the video features with the text embeddings.

At each diffusion step, attention scores are computed as:

\begin{equation}
    \mathbf{A} = \text{softmax} \left( \frac{\mathbf{Q} \mathbf{K}^\top}{\sqrt{d_k}} \right),
\end{equation}

where $\mathbf{Q} = \mathbf{X} \mathbf{W}_Q \in \mathbb{R}^{T \times H \times W \times d_k}$ are the query vectors obtained from the video features, and $\mathbf{K} = \mathbf{E} \mathbf{W}_K \in \mathbb{R}^{L \times d_k}$ are the key vectors from the text embeddings. The attention map $\mathbf{A} \in \mathbb{R}^{(T \times H \times W) \times L}$ highlights the relevance of each word in the text prompt to the spatiotemporal video features. The output of the cross-attention layer is then:
$\mathbf{X}' = \mathbf{A} \mathbf{V},$
where $\mathbf{V} = \mathbf{E} \mathbf{W}_V \in \mathbb{R}^{L \times d_v}$ are the value vectors from the text embeddings. The updated video features $\mathbf{X}' \in \mathbb{R}^{T \times H \times W \times d_v}$ are fused with the text semantics, allowing the model to condition the visual generation on the input prompt. This cross-attention mechanism enables alignment between the spatiotemporal dynamics of the video and the linguistic cues, ensuring that the generated video is semantically consistent with the text in the diffusion process.\\

\noindent \textbf{Noise frequency vs Objects.} Video diffusion models often exhibit temporal inconsistency due to a mismatch between the temporally correlated noise used in training and the independent Gaussian noise employed during inference, degrading the quality of generated videos \cite{wu2023freeinit}. To address this, the initial noise $z_T$ can be refined during inference. The process begins with standard DDIM sampling \cite{ho2020denoising} with diffusion time steps $T$ to $0$, to generate a clean video latent \( z_0 \), followed by forward diffusion steps to obtain a noisy latent \( z_T \), which preserves improved temporal consistency. The key concept is the noise reinitialization process, where spatio-temporal frequency filtering preserves the low-frequency components of \( z_T \) while replacing the high-frequency components with newly sampled Gaussian noise $\eta$ \cite{wu2023freeinit}:
\begin{align}
F^{low}_{z_T} = \text{FFT3D}(z_T) \odot \mathcal{H}, \\
F^{high}_{\eta} = \text{FFT3D}(\eta) \odot (1 - \mathcal{H}), \\
z'_T = \text{IFFT3D}(F^{low}_{z_T} + F^{high}_{\eta}),
\end{align}
Here, \( \mathcal{H} \) is a spatio-temporal low-pass filter, and \text{FFT3D} and \text{IFFT3D} represent the 3D Fast Fourier Transform and its inverse, respectively. The reinitialized noise \( z'_T \) is then used for subsequent DDIM sampling iterations, improving temporal consistency and video quality. It is observed that low-frequency signals are more resilient to additive noise, leading the diffusion model to naturally retain layout and shape information from the initial noise, which strongly align with the foreground object \cite{wu2023freeinit,qiu2024freetraj}.


\begin{figure*}[htbp]
\centering
\includegraphics[width=0.8\linewidth]
{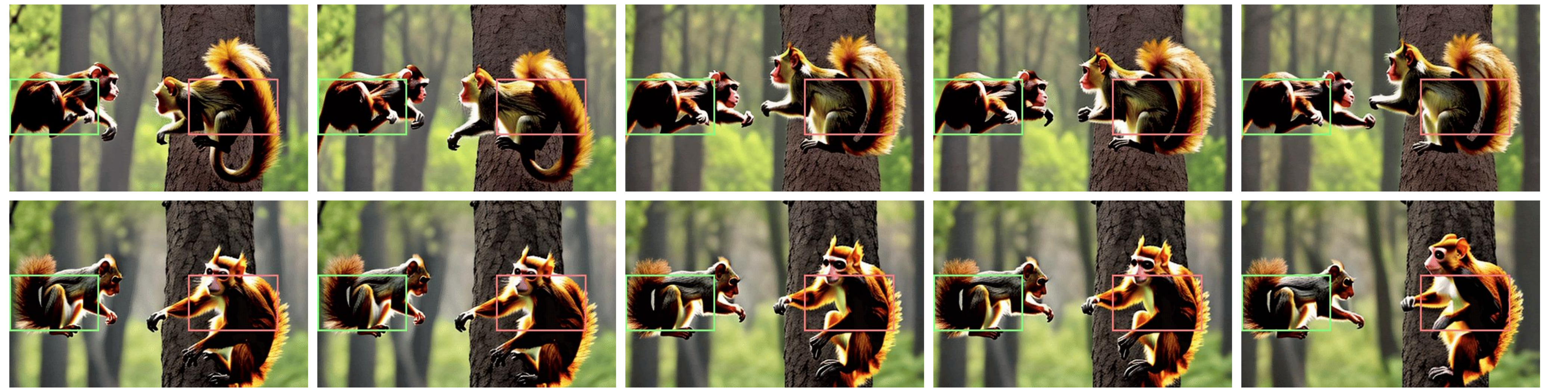}
\vspace{-10pt}
\caption{Video frames generated by the prompt ``A monkey and a squirrel on a tree." \textbf{Top row shows the results without attention re-weighting}, where features are mixed—e.g., the squirrel has the face of a monkey, and the monkey has a squirrel's tail. \textbf{Bottom row shows the improved results after applying attention re-weighting}, where both the monkey and squirrel retain their correct, distinct features.
}

\label{Fig:attn-re}
\end{figure*}

\subsection{Trajectory Generation}
By restructuring low-frequency noise components across frames, we can control inter-frame content correlation, effectively using low-frequency components guiding object movements in the generated video. To manipulate the low-frequency component across frames for object motion control, we aim for a purely text-conditioned approach, enabling flexible, open-domain generation without the need for predefined object trajectories. 
We leverage the open-world knowledge of large language models (LLMs), and use an LLM as the scene director, responsible for planning motion and interactions that align with real-world physics and the specified prompt. Specifically, we generate object trajectories using the LLM by giving in-context learning with a set of few-shot examples containing prompts, objects, and their corresponding trajectories, represented as bounding boxes over time. These trajectories, denoted as a sequence of $(x_t, y_t)$ over time $t$, describe the positions of objects in the scene in each frame. Let the prompt be represented as $\mathbf{p}$ and the set of objects as $\mathcal{O} = \{ o_1, o_2, \dots, o_n \}$, where $n$ is the number of objects in the scene. For each object $o_i$, its trajectory over time is represented as a sequence of bounding boxes $\mathcal{T}_i = \{ (x_{i,t}, y_{i,t}) \}_{t=1}^T$, where $T$ is the number of frames, and $(x_{i,t}, y_{i,t})$ represents the position of object $o_i$ at time $t$.
Referring to the relationship between the prompts and the corresponding motion patterns in the $m$ trajectory examples $\{ (\mathbf{p}_j, \mathcal{T}_j) \}_{j=1}^m$, the LLM imagines the scene for a new prompt $\mathbf{p}'$, and predicts the trajectories $\hat{\mathcal{T}}_i$ for each object $o_i$:
\begin{equation}
    \hat{\mathcal{T}}_i = \text{LLM}(\mathbf{p}', o_i, \{ (\mathbf{p}_j, \mathcal{T}_j) \}_{j=1}^m).
\end{equation}

This produces $n$ object trajectories $\{ \hat{\mathcal{T}}_i \}_{i=1}^n$, each represented as a sequence of bounding boxes $\hat{\mathcal{T}}_i = \{ (\hat{x}_{i,t}, \hat{y}_{i,t}) \}_{t=1}^T$ object $o_i$ over time.

\subsection{Multi-Object Generation}
In the multi-object generation phase, we link the initial noise of each object with its LLM predicted trajectory \( \{ \hat{\mathcal{T}}_i \}_{i=1}^n \). Starting from these generated object trajectories, the primary goal is to propagate noise accordingly across frames, establishing a ``noise flow" that defines each object's movement. For each video, noise \( \epsilon \) is sampled only for the first frame, serving as the initial noise. This initial noise is then shifted based on each object's trajectory \( z_T^{\text{flow}} \) to achieve consistent noise alignments across the sequence of frames. The noise at each position \([i, j]\) for frame \( f \) is updated as \cite{qiu2024freetraj}:
\begin{equation}
 \epsilon[i, j]^f = \epsilon[(i - \Delta_i) \bmod H, (j - \Delta_j) \bmod W]^{f-1},
\end{equation}

where \(\Delta_i\) and \(\Delta_j\) are the object-specific shifts determined by the predicted trajectory in \( z_T^{\text{flow}} \). This approach allows for noise propagation across the frames, aligning both foreground and background movements in accordance with each object’s designated trajectory. To control noise within a specific region for each object, an input mask is applied for a localized noise flow. Starting with an initial random local noise \( \epsilon_{\text{local}} \), sampled specifically within the masked area, we generate \( F \) frames of independent random noise sequences \([ \epsilon_1, \epsilon_2, \ldots, \epsilon_F ]\) for each object trajectory. For each frame \( f \), the noise \( \epsilon_f \) within the input mask is replaced by \( \epsilon_{\text{local}} \) as follows:

\begin{equation}
\tilde{\epsilon}_f[i, j] = 
\begin{cases} 
\epsilon_f[i, j] & \text{if } M_f[i, j] = 0, \\ 
\epsilon_{\text{local}}[i^*, j^*] & \text{if } M_f[i, j] = 1,
\end{cases}    
\end{equation}

where \( \epsilon_f \) and \( \epsilon_{\text{local}} \sim \mathcal{N}(0, I) \). Here, \( M_f \) is the input mask for frame \( f \), where \( M_f[i, j] = 1 \) indicates that position \( (i, j) \) lies within the object’s bounding box, and \( M_f[i, j] = 0 \) otherwise. The indices \( (i^*, j^*) \) represent the relative position within the bounding box of each object, ensuring that the noise stays within the local area of the trajectory.

By leveraging the diffusion model’s learned priors on single object generation and movement, the trajectory-specific noise flows allow each object to follow its designated path in accordance with the prompt. For example, given a prompt such as ``A frog is jumping and a rat is walking", the LLM generates trajectory patterns—such as a jumping motion for one bounding box and a walking motion for another—while the diffusion model uses its prior knowledge to place each object accordingly on its path and generates the corresponding appearance.


\begin{figure*}[htbp]
\centering
\includegraphics[width=.9\linewidth]
{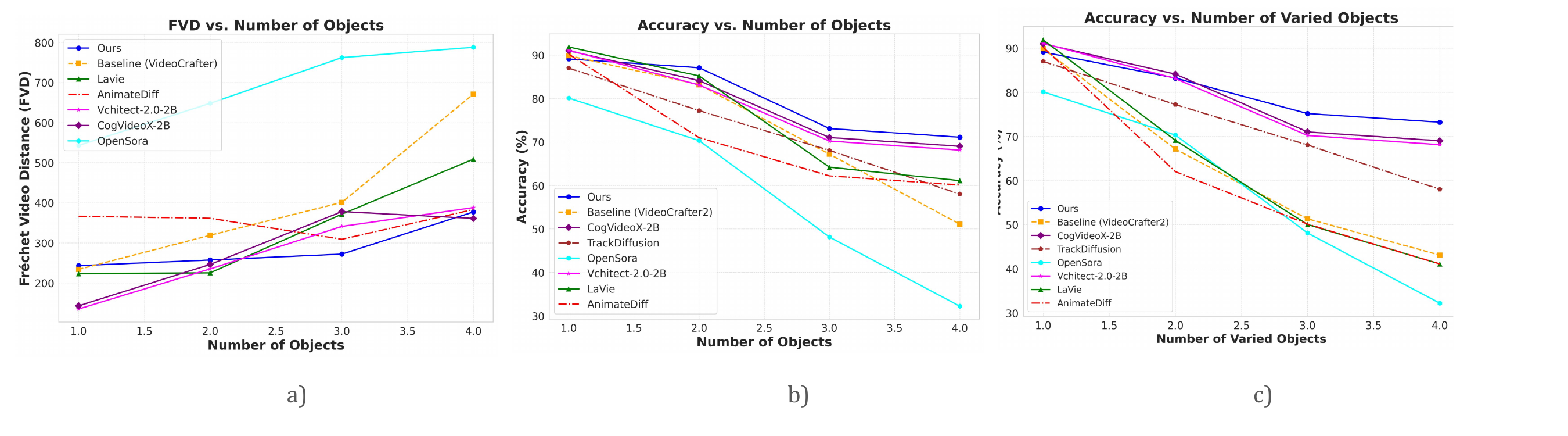}
\vspace{-20pt}
\caption{Multi-object generation quality and accuracy. We report a) FVD and b) object generation accuracy as the number of objects increases and c) varied objects. Please refer to \cref{sec:results} for details.
}

\label{Fig:obj}
\end{figure*}

\subsection{Multi-Object Refinement}

Although noise re-initialization can help guide the trajectories of multiple objects, challenges, such as competing objects merging or morphing into each other, remain. 
For more precise control over the appearance and interactions of the objects in the scene, we further leverage the LLM-predicted trajectory and apply fine-grained control to the influence of each text token on the generated video. 
Specifically, given the predicted trajectories, we reweight the attention scores along the trajectory of each object according to its designated text tokens, allowing selective strengthening or weakening of each text token's contribution to the video features.

Consider a prompt \( \mathbf{P} = \text{``a cat and a dog playing"} \), and let the objects \( \mathcal{O} = \{ o_1, o_2 \} \) represent the cat and the dog, respectively. The goal is to prevent the influence of the ``dog" text token on the visual tokens within the bounding box region assigned to the cat by manipulating the corresponding attention map. Similarly, the influence of the ``cat" text token should be suppressed in the attention map for the bounding box region assigned to the dog. This refinement ensures that each object retains its identity and movement pattern without being influenced by the other objects.

For each text token \( j \) in the prompt, we associate an attention map \( \mathcal{A}_j \), which determines the extent to which the text token affects the resulting video. To selectively control the influence of a text token \( j^* \) on a particular object \( o_i \) within its bounding box region \( \mathcal{B}_i \), we introduce a scaling parameter \( c \in [-2, 2] \) that adjusts the attention map specifically for that text token within the bounding box region. This results in an updated attention map \( \hat{\mathcal{A}}_{j^*}^{\mathcal{B}_i} \), which scales the influence of the text token \( j^* \) by the factor \( c \) only within \( \mathcal{B}_i \):
\begin{equation}  \hat{\mathcal{A}}_{j^*}^{\mathcal{B}_i} = c \cdot \mathcal{A}_{j^*}^{\mathcal{B}_i}
\end{equation}

The scaling factor \( c \) can be used to either amplify (\( c > 1 \)) or attenuate (\( c < 1 \)) the effect of the text token \( j^* \) specifically within the object’s bounding box region \( \mathcal{B}_i \). For example, by setting \( c = 0 \), we completely remove the influence of the text token \( j^* \) within the bounding box region \( \mathcal{B}_i \).

Let \( \mathbf{P} \) be the prompt, and let \( \{ \mathcal{A}_j \}_{j=1}^k \) be the set of attention maps corresponding to the \( k \) text tokens in the prompt. The final attention map for object \( o_i \) after refinement is given by:
\begin{equation}
 \hat{\mathcal{A}}_i = \sum_{j=1}^k \alpha_j \cdot \mathcal{A}_j
\end{equation}

where \( \alpha_j = c \) if \( j = j^* \) (i.e., the text token whose influence is adjusted within the bounding box region of \( o_i \)), and \( \alpha_j = 1 \) for all other text tokens. This formulation allows for fine-grained control over the spatial regions influenced by each text token in the prompt, effectively adjusting the impact of specific text tokens within the object’s designated region and prevent cross-object interference.

By selectively re-weighting the attention maps, we ensure that objects are influenced only by their assigned text tokens. In the example where ``a cat and a dog playing" is the prompt, the attention map corresponding to the text token ``dog" is attenuated or removed in regions corresponding to the bounding boxes of the ``cat". Conversely, the attention map for ``cat" is modified similarly for the ``dog" bounding boxes. This process ensures that the generated trajectories and appearances of each object in the video remain independent and coherent, based on the prompt's semantic structure. An example is shown in Fig. \ref{Fig:attn-re}.

\section{Experiments and Results}
\subsection{Experimental Setup}
We perform DDIM sampling \cite{song2020denoising} with 50 steps and deterministic sampling. The video resolution is fixed at 320 × 512 pixels with a length of 16 frames. All inference is done on an AMD MI250 GPU with 128GB RAM using PyTorch.

\subsection{Baselines} 
There is no prior work explicitly focused on multi-object video generation. Therefore, we compare the multi-object generation capabilities of state-of-the-art open-source video generation models. For specifically multi-object scenarios, we include TrackDiffusion \cite{li2023trackdiffusion}, which is trained on object instances and their trajectories. Beyond this, we evaluate two broader categories of models: text-to-video (T2V) models, including VideoCrafter2 \cite{chen2024videocrafter2}, AnimateDiff \cite{guo2023animatediff}, OpenSora \cite{opensora}, CogVideoX-2B \cite{yang2024cogvideox}, and Vchitect-2.0-2B \cite{chen2024videocrafter2}; and image-to-video (I2V) models, which generate videos from an initial image containing multiple objects, including Stable Diffusion \cite{blattmann2023stable}, DynamiCrafter \cite{xing2024dynamicrafter}, VideoCrafter \cite{chen2024videocrafter2}, and I2V-Gen-XL \cite{zhang2023i2vgen}. Additionally, we evaluate MOVi against commercial models, including Luma AI~\cite{lumaAI} and Gen-2~\cite{runwayml_gen2}, to provide a comprehensive comparison.


\subsection{Main Results} 
\label{sec:results}


\begin{figure*}[htbp]
\centering
\includegraphics[width=1\linewidth]
{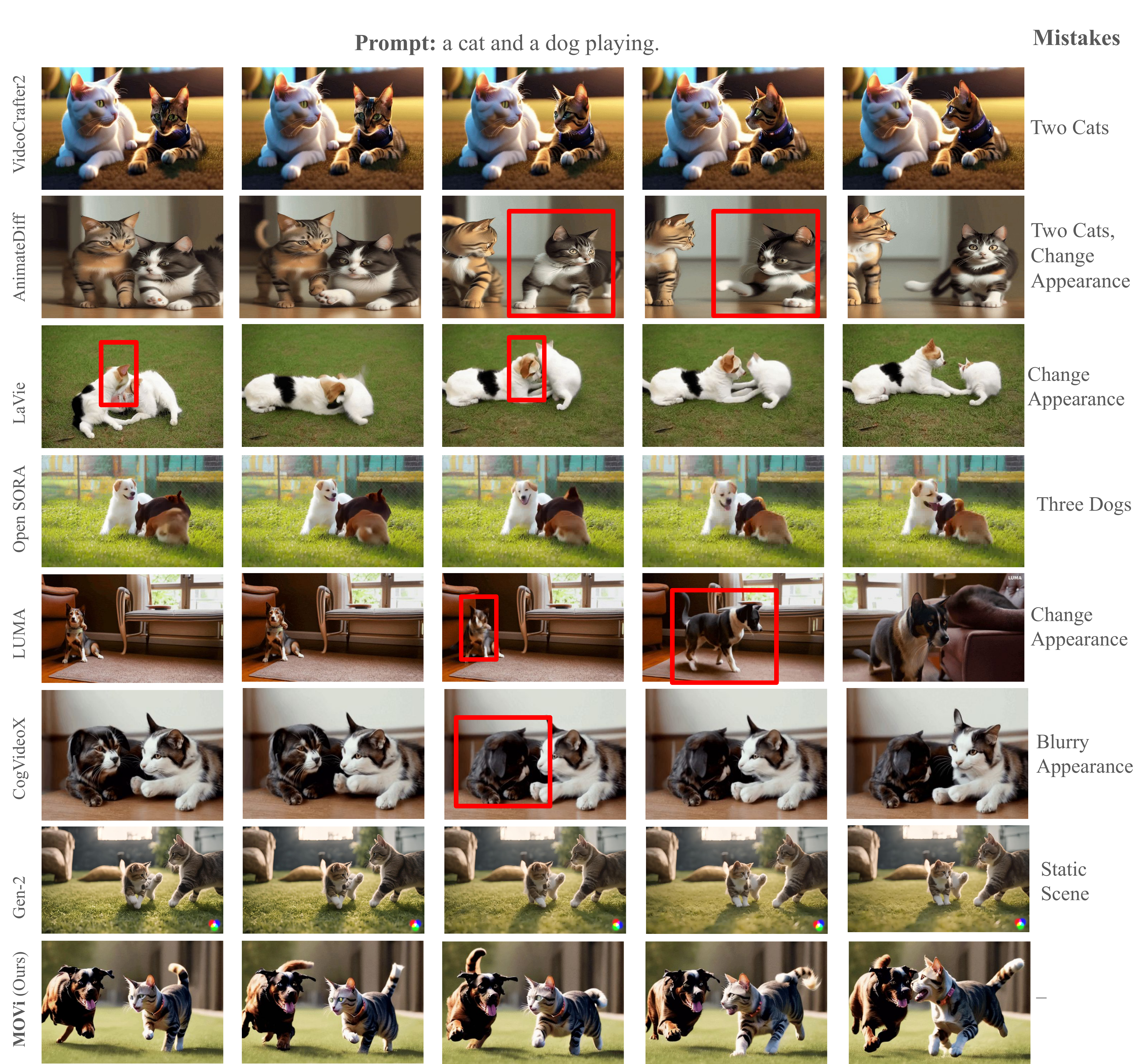}
\vspace{-5pt}
\caption{Qualitative comparison between MOVi and state-of-the-art T2V models given the prompt ``a cat and a dog playing." Notably, existing methods struggle to generate two distinct objects, often producing either only cats or a combined and indistinct shape of both animals. In contrast, MOVi successfully generates both the cat and the dog as separate and well-defined objects.
}
\label{fig:quals}
\vspace{-20pt}
\end{figure*}
\vspace{1mm}
\noindent \textbf{Multi-Object Generation Quality and Accuracy.} We design a set of GPT-generated prompts tailored for different numbers of YOLO \cite{zhang2022real} object classes and use these prompts to generate corresponding videos. For each video, we evaluate object classification and counting accuracy using YOLO-v5 \cite{zhang2022real}, and report the average performance. \cref{Fig:obj} illustrates the multi-object generation capabilities of MOVi compared to VideoCrafter2 \cite{chen2024videocrafter2}, LaVie~\cite{wang2023lavie}, OpenSora \cite{opensora}, CogVideoX-2B \cite{yang2024cogvideox}, Vchitect-2.0-2B \cite{chen2024videocrafter2}, TrackDiffusion \cite{li2023trackdiffusion}  and AnimateDiff~\cite{guo2023animatediff}. In~\cref{Fig:obj} (a), we plot the Fréchet Video Distance (FVD)~\cite{unterthiner2019fvd} as a function of the number of generated objects. Our model maintains FVD when generating up to two objects, but significantly outperforms the baseline as well as LaVie~\cite{wang2023lavie} and AnimateDiff~\cite{guo2023animatediff} as the object count increases, demonstrating its robustness in complex scenes. \cref{Fig:obj} (b) presents a comparison of detection accuracy when generating multiple instances of the same object (\eg, two dogs, and three lizards). As the number of objects increases, the baseline model and LaVie~\cite{wang2023lavie} and AnimateDiff~\cite{guo2023animatediff} experiences a steep decline in accuracy, while our model retains higher performance. This trend persists even with varied objects in the prompt (\eg, different species) as in~\cref{Fig:obj} (c), further highlighting the scalability and precision of our approach in handling more complex, multi-object scenarios.

\noindent \textbf{Comparison with Baseline Models.} Table~\ref{tab:multi_object_comparison} provides a detailed comparison of various text-to-video (T2V) and image-to-video (I2V) models across key motion and fidelity metrics, including Dynamic Degree, Accuracy, Motion Smoothness, Weighted Average, and Frechet Video Distance (FVD). These metrics collectively assess whether a generated video exhibits sufficient motion, accurately represents the prompt-specified objects, and strikes a balance between object accuracy and motion quality. This is crucial, as some models may generate highly dynamic videos with incorrect objects, while others may produce accurate objects with limited motion. Our model, MOVi [UNet-base], achieves a Dynamic Degree of 100\%, significantly outperforming all other T2V models due to its trajectory-based motion generation. Additionally, MOVi attains the highest object Accuracy among evaluated models. The weighted average metric, which balances motion and accuracy, reaches 0.85, demonstrating a substantial improvement over competing approaches. MOVi also maintains near-perfect Motion Smoothness, comparable to leading I2V models. While OpenSora and AnimateDiff achieve high Dynamic Degrees, their elevated FVD scores indicate lower perceptual quality. In contrast, MOVi substantially reduces FVD, highlighting its ability to generate visually coherent and dynamic videos with accurate object representation.


\begin{table*}[t]
    \centering
    \caption{Comparison with baseline models. Metrics include Dynamic Degree, Accuracy, Motion Smoothness, a weighted average of Dynamic Degree and Accuracy, and FVD. T2V and I2V subsequently refer to text-to-video and image-to-video generation models.}
    \resizebox{0.8\textwidth}{!}{%
    \begin{tabular}{lccccc c}
        \toprule
        \textbf{Model Name} & \textbf{Model Type} & \textbf{Dynamic Degree} & \textbf{Accuracy} & \textbf{Weighted Average} & \textbf{Motion Smoothness} & \textbf{FVD $\downarrow$} \\
        \midrule
        TrackDiffusion \cite{li2023trackdiffusion} & T2V & 71\% & 0.35 & 0.53 & 0.95 & 531.2 \\
        VideoCrafter2 \cite{chen2024videocrafter2} & T2V & 45\% & 0.42 & 0.43 & 0.96 & 321 \\
        AnimateDiff \cite{guo2023animatediff} & T2V & 17\% & 0.39 & 0.28 & 0.98 & 704 \\
        OpenSora \cite{opensora} & T2V & 93\% & 0.42 & 0.67 & 0.91 & 731 \\
        CogVideoX-2B \cite{yang2024cogvideox} & T2V & 64\% & 0.59 & 0.61 & 0.97 & \textbf{241} \\
        Vchitect-2.0-2B \cite{chen2024videocrafter2} & T2V & 58\% & 0.70 & 0.64 & 0.97 & 233.7 \\
        \midrule
        Stable Diffusion \cite{blattmann2023stable} & I2V & 52\% & 0.69 & 0.61 & 0.98 & 519.1 \\
        DynamiCrafter \cite{xing2024dynamicrafter} & I2V & 42\% & 0.69 & 0.55 & 0.97 & 258.3 \\
        VideoCrafter \cite{chen2024videocrafter2} & I2V & 21\% & 0.53 & 0.37 & 0.98 & 510 \\
        I2V-Gen-XL \cite{zhang2023i2vgen} & I2V & 23\% & 0.43 & 0.33 & 0.98 & 550.3 \\
        \midrule                               
        MOVi \cite{chen2024videocrafter2} (Ours) & T2V & \textbf{100\%} & \textbf{0.71} & \textbf{0.85} & \textbf{0.98} & 295.3
        \\
        \bottomrule
    \end{tabular}}
    \label{tab:multi_object_comparison}
    \vspace{-15pt}
\end{table*}

        
        

\begin{table}[t]
    \centering
    \caption{Comparison with commercial models. We report ``Dynamic Degree" from VBench~\cite{huang2024vbench}, and motion and object ratings from human preference study. ($^*$) indicates a commercial model.}
    \resizebox{\columnwidth}{!}{\begin{tabular}{l|ccc}
        \toprule
        \textbf{Model Name} & \textbf{Dynamic Degree} & \textbf{Motion Rating} & \textbf{Object Rating} \\
        \midrule

        Gen-2$^*$ \cite{runwayml_gen2}  &                      15\% & 1.86   & 2.60  \\
        Luma AI$^*$~\cite{lumaAI}                & 33\% & 1.95  & \textbf{4.47}  \\
        VideoCrafter2~\cite{chen2024videocrafter2} & 48\% & 2.47  & 2.34  \\
        \textbf{MOVi (Ours)}      & \textbf{100\%} & \textbf{4.52}  & 3.47  \\
        \bottomrule
    \end{tabular}}
    \label{multi_object}
    \vspace{-15pt}
\end{table}

\vspace{1mm}
\noindent \textbf{Qualitative Results.} Existing metrics for multi-object generation often fall short, as subtle feature mix-ups between competing objects may not significantly impact object accuracy scores. We present qualitative results in \cref{fig:quals}, comparing our method with other state-of-the-art open-source T2V models~\cite{wang2023lavie, chen2024videocrafter2, guo2023animatediff,opensora, yang2024cogvideox}, and commercial models~\cite{runwayml_gen2,lumaAI}. Unlike baseline approaches that generate merged objects, frequently mix features between objects, or exhibit minimal motion, MOVi leverages trajectory guidance and attention refinement to achieve visually compelling results.

\begin{figure}[htbp]
\centering
\includegraphics[width=1\linewidth]
{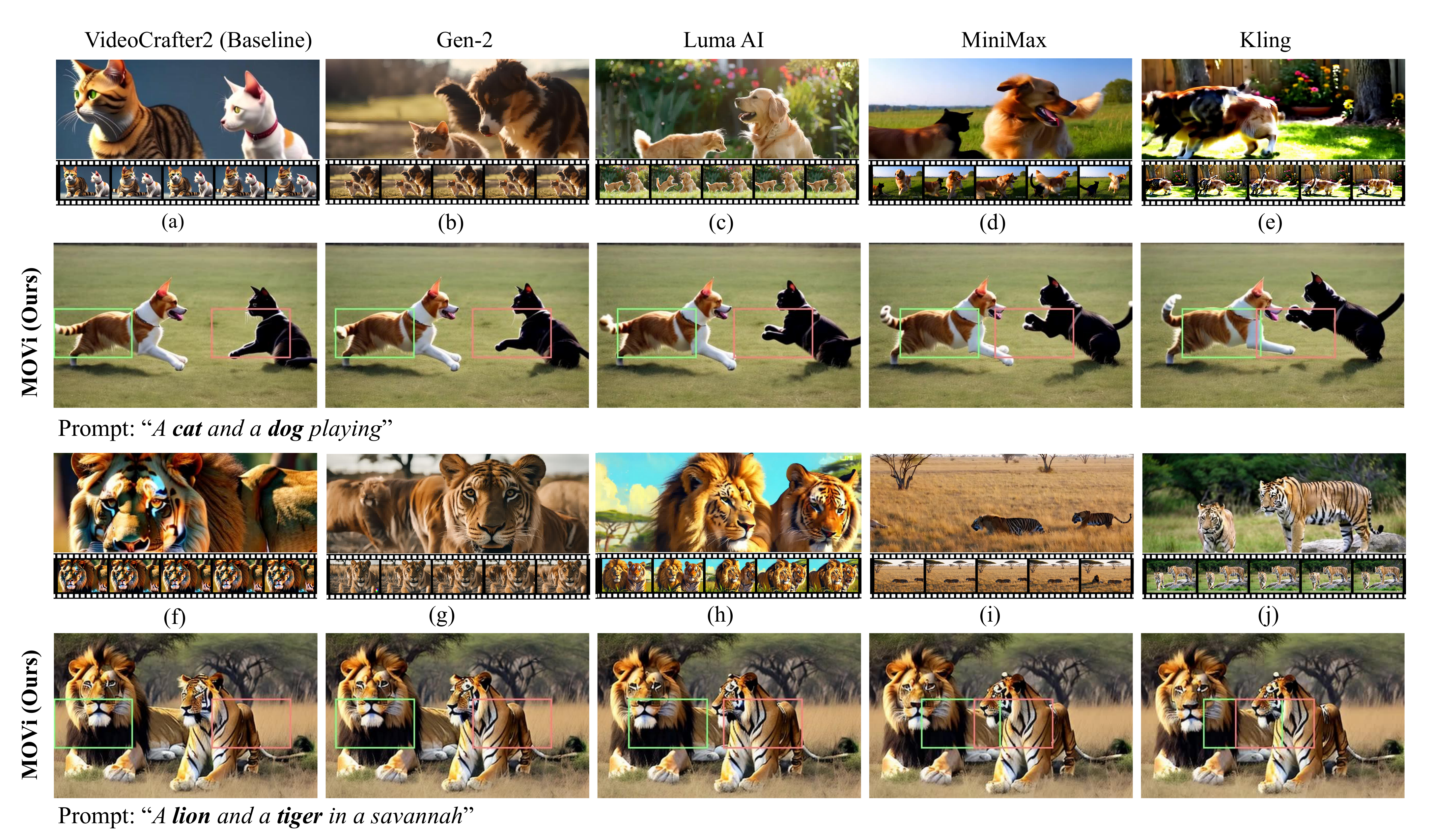}
\vspace{-15pt}
\caption{Results of multiple-object video generation. Baseline~\cite{chen2024videocrafter2} and commercial models~\cite{runwayml_gen2, lumaAI, minimax,kling} often struggle to generate multiple distinct objects as specified in the prompt, such as failing to generate the correct number (f, g) and categories (a, i, j) of objects, mixing features across different objects (c, d, e, h), and failing to capture object interactions and limiting the object movements (b).
}
\label{fig:quals_com}
\vspace{-15pt}
\end{figure}
\noindent \textbf{Comparison with Commercial Models.} 
We compare our approach with two commercial models: Luma \cite{lumaAI} and Gen-2 \cite{runwayml_gen2}. We include Gen-2 because it performs better than Gen-3~\cite{runwayml_gen3} in VBench's multi-object metric. Commercial models can sometimes generate multiple objects, but we observe that their motion dynamics are limited. To quantify this, we assess the degree of dynamic motions, drawing on VBench's ``dynamic degree" metric, which uses RAFT \cite{teed2020raft} to evaluate the extent of motion in synthesized videos. This metric is crucial because a completely static video could still perform well in the multi-object category, underscoring the need to assess whether models generate substantial object movement. The results are shown in~\Cref{multi_object}. We achieve a dynamic degree of 100\%, significantly outperforming the baseline models, which indicates that our method can generate multi-object with dynamic motions. Figure \ref{fig:quals_com} shows visual comparison with additional commercial models.

   

\vspace{1mm}
\noindent \textbf{Human Preference Study.}
To further evaluate the motion and object generation quality of the generated videos, we present a human preference study. We compare our model with the baseline model, VideoCrafter2 \cite{chen2024videocrafter2}, as well as two commercial models, Gen-2 \cite{runwayml_gen2} and LumaAI \cite{lumaAI}. A group of 23 participants evaluated and rated the videos based on alignment with the provided text prompts. Participants were asked to rate the accuracy of the objects and their motions, as described in the prompt, on a scale of 1 to 5. The results are summarized in Table \ref{multi_object}. We achieved the highest motion rating and the second highest object rating, demonstrating the high quality of our generated multi-object videos even compared with commercial models.

\vspace{1mm}
\noindent \textbf{Additional Results.} Evaluating multi-object scenarios presents challenges. While the multi-object metric from VBench \cite{huang2024vbench} is considered, it is limited to two objects and does not regard motion of the objects. To provide a broader evaluation, we incorporate Fréchet Video Distance (FVD) \cite{unterthiner2019fvd} on UCF-101 \cite{soomro2012ucf101} dataset and CLIP similarity score~\cite{hessel2021clipscore} for MSR-VTT~\cite{xu2016msr}, enabling comparisons with other models in the absence of a standardized video benchmarking dataset. The results are presented in the Appendix. 

\subsection{Ablation Study}
\vspace{1mm}
\noindent \textbf{Effect of Proposed Enhancements.} We study the effects of different components of MOVi in Table \ref{abl}. First, trajectory insertion, which adds object trajectories based on the number of objects specified in the prompt, significantly improves the object class generation accuracy for up to four objects. Additionally, attention refinement further enhances performance by refining the semantics of each object, leading to more accurate and coherent video outputs.

\begin{table}[]
	\centering
		\caption{
Ablation study on key design components. 
  }

	\resizebox{0.8\columnwidth}{!}{
		\begin{tabular}{c|ccc}  
			\Xhline{3\arrayrulewidth} 
			  \multirow{2}{*}{\color[HTML]{000000}\textbf{ Model Name}} & \multicolumn{3}{c}{\color[HTML]{000000} \textbf{Accuracy}} \\ 
			& {\color[HTML]{000000} \textbf{n=2}} & {\color[HTML]{000000} \textbf{n=3}} & {\color[HTML]{000000} \textbf{n=4}} \\ \Xhline{2\arrayrulewidth} 
			Baseline \cite{chen2024videocrafter2} & 0.824 & 0.692 & 0.53 \\ 
			{\color[HTML]{000000} Trajectory Guidance } & {\color[HTML]{000000} 0.849} & {\color[HTML]{000000} \textbf{0.739}} & {\color[HTML]{000000} 0.695} \\
			\makecell{Trajectory Guidance \\ + Attention Refinement} & \textbf{0.861} & 0.731 & \textbf{0.71} \\ 
			\Xhline{3\arrayrulewidth} 
		\end{tabular}
	}

	\label{abl}

\end{table}

\vspace{1mm}
\noindent \textbf{Choosing Trajectory Generator.} We experimented with various models for trajectory generation, including multimodal LLMs such as CogVLM~\cite{wang2023cogvlm} and LLaVA~\cite{liu2024visual}, both fine-tuned on the YouTube VIS dataset~\cite{yang2019video}, and LLMs such as Llama 3.1 8B and 405B~\cite{touvron2023llama} without any finetuning. To evaluate the generated trajectories, which inherently lack a clear ground truth due to the possibility of multiple plausible outcomes for a given prompt, we employ a two-step approach. First, we generate videos from the predicted trajectories and assess their quality FVD. Viable trajectories should yield lower FVD scores, indicating more realistic and coherent videos. Second, we utilize GPT-4's~\cite{achiam2023gpt} understanding of motion realism, physical plausibility, smoothness, and consistency to rate each trajectory on a scale from 0 to 10. The results are presented in Table \ref{fvds}. Llama-3.1, especially the 405B model, with its zero-shot capability and broader world understanding, outperformed the fine-tuned models in generating more versatile trajectories. Therefore we selected the Llama 3.1 405B model for this task.
\begin{table}[]
	\centering
		\caption{
Comparison of quantitative results in terms of FVD and GPT-4 Evaluation. The best results are in \textbf{Bold}. ``FT'' means the model is fine-tuned on prompts and their trajectories. 
  }

	\resizebox{0.9\columnwidth}{!}{
		\begin{tabular}{l|cc}  
                \Xhline{3\arrayrulewidth} 

			{\color[HTML]{000000} \textbf{Model Name}} & {\color[HTML]{000000} \textbf{FVD ($\downarrow$})} & {\color[HTML]{000000} \textbf{GPT-4 Evaluation ($\uparrow$)}} \\ \Xhline{2\arrayrulewidth} 
			CogVLM-FT \cite{wang2023cogvlm} & 795.01
 & 6.3 \\ 
			{\color[HTML]{000000} LLaVa-FT \cite{liu2024visual}} & {\color[HTML]{000000} 1093} & 3.3 \\
			LLaVa-1.5-FT \cite{liu2024visual} & 832.3& 3.9 \\ 
   
		Llama 3.1 8B \cite{touvron2023llama} & 242.6 & 8.3 \\ 
  
		\textbf{Llama 3.1 405B} \cite{touvron2023llama} & \textbf{208.1}
& \textbf{9.3} \\ 
		\Xhline{3\arrayrulewidth} 

		\end{tabular}
	}

	\label{fvds}
\vspace{-20pt}
\end{table}




\section{Conclusion}
\vspace{-1mm}

In this paper, we introduce MOVi, a training-free method that improves the multi-object video generation capabilities of video diffusion models. We leverage the open-world knowledge of LLMs to plan the object trajectories given a text prompt, and apply the trajectories to video generation through noise re-initialization. Object appearance and interactions are further refined by attention score reweighting according to their trajectories. Experiment results demonstrate significantly improved multi-object video generation, surpassing even commercial models.


\section{Supplementary}

\section{Additional Results}

\noindent \textbf{VBench Multi-Object Score.} VBench \cite{huang2024vbench} evaluates multiple object compositionality by using GRiT \cite{wu2025grit} for frame-wise object detection, ensuring all prompt-specified objects appear simultaneously in each frame. The success rate is reported as the proportion of frames where all required objects are detected. In the evaluation prompt, COCO objects are grouped into logical categories, such as animals, indoor items, dining objects, bathroom items, and outdoor items. The VBench results are reported in Table \ref{vbench}. MOVi significantly boosts the multi-object score of the base VideoCrafter2 model from 40.66\% to 62.19\%, outperforming commercial models such as Gen-2~\cite{runwayml_gen2} and Gen-3~\cite{runwayml_gen3}. Our result is further improved to 71.95\% by applying rejection sampling during video generation, outperforming the second best model Vchitect-2.0-2B~\cite{vchitect} in the VBench multi-object score leaderboard. 

\begin{table}[t]
    \centering
    \caption{Multi-object score on VBench~\cite{huang2024vbench}. ($^*$) indicates commercial models. ``re'' stands for rejection sampling.}
    \resizebox{0.75\columnwidth}{!}{\begin{tabular}{l|l}
        \toprule
        \textbf{Model Name} & \textbf{Multi-Object Score} \\
        \midrule
        MiniMax-Video-01$^*$~\cite{minimax}                          & 76.04\%  \\
        Vchitect-2.0-2B~\cite{vchitect}                           & 69.35\%  \\
        Kling$^*$~\cite{kling}      & 68.05\%  \\
        CogVideoX-2B~\cite{yang2024cogvideox}        & 62.63\%  \\
                OpenSora (v1.2) \cite{opensora} & 58.41\% \\
                Gen-2*~\cite{runwayml_gen2}       & 55.47\%  \\
                Gen-3*~\cite{runwayml_gen3}       & 53.64\%  \\
                        Emu-3~\cite{wang2024emu3}       & 44.64\%  \\
     AnimateDiff (v2) \cite{guo2023animatediff} & 36.88\% \\
        LaVie \cite{wang2023lavie} & 33.32\% \\ \hline
        
VideoCrafter2 ~\cite{chen2024videocrafter2} & 40.66\%\\
        \textbf{MOVi (Ours)}        & 62.19\% (\textbf{+21.53\%}) \\
        
 \textbf{MOVi-re (Ours)} & 71.95\% (\textbf{+31.29\%})\\
        \bottomrule
    \end{tabular}}
    \label{vbench}
\end{table}

\subsection{Results on UCF-101 and MSR-VTT}
 To provide a broader evaluation, we incorporate Fréchet Video Distance (FVD) \cite{unterthiner2019fvd} on UCF-101 \cite{soomro2012ucf101} dataset and CLIP similarity score~\cite{hessel2021clipscore} for MSR-VTT~\cite{xu2016msr}, enabling comparisons with other models in the absence of a standardized video benchmarking dataset. The results are presented in Table \ref{fvds} and Table \ref{quan}.
\begin{table}[!h]
    \centering
    \caption{Quantitative zero-shot comparisons on UCF101 \cite{soomro2012ucf101}. ``$\downarrow$" denotes the lower the better. ``$\uparrow$" denotes the higher the better. ``--" means the number is not reported in the original paper.}
    \begin{tabular}{llc}
        \toprule
        Method & Resolution & FVD $\downarrow$ \\
        \midrule
        LVDM \cite{he2022latent} & 16 $\times$ 128 $\times$ 128 & 372 \\
        MagicVideo \cite{zhou2022magicvideo} & 16 $\times$ 256 $\times$ 256  & 655 \\
        CogVideo \cite{hong2022cogvideo} & 16 $\times$ 128 $\times$ 128 & 626 \\
        VideoLDM \cite{blattmann2023align} & 16 $\times$  -- $\times$  -- & 550.61 \\ 
                Show-1 \cite{zhang2024show} & - $\times$  -- $\times$  -- & 394.46 \\
        \textbf{MOVi (Ours}) & 16 $\times$ 256 $\times$ 256 & \textbf{371} \\
        \bottomrule
    \end{tabular}
    \label{fvds}
\end{table}

\begin{table}[!h]
    \centering
    \vskip -10pt
    \caption{Quantitative zero-shot comparisons on MSRVTT \cite{soomro2012ucf101}. ``$\downarrow$" denotes the lower the better. ``$\uparrow$" denotes the higher the better. ``--" means the number is not reported in the original paper.}
    \begin{tabular}{llc}
        \toprule
        Method & CLIPSIM $\uparrow$ & FID $\downarrow$ \\
        \midrule
        LaVie \cite{wang2023lavie} & 0.2949 & - \\
        ModelScope \cite{wang2023modelscope} & 0.293  & \textbf{11.09} \\
        CogVideo \cite{hong2022cogvideo} & 0.2631 & 23.59 \\
        VideoLDM \cite{blattmann2023align} & 0.2929 & - \\
             Show-1 \cite{zhang2024show} & \textbf{0.3072} & 13.08 \\
    
        \textbf{MOVi (Ours}) & \textbf{0.3072} & 22.13 \\
        \bottomrule
    \end{tabular}
    \label{quan}
\end{table}

\subsection{More Qualitative Results} 
\begin{figure*}[htbp]
\centering
\includegraphics[width=1\linewidth]
{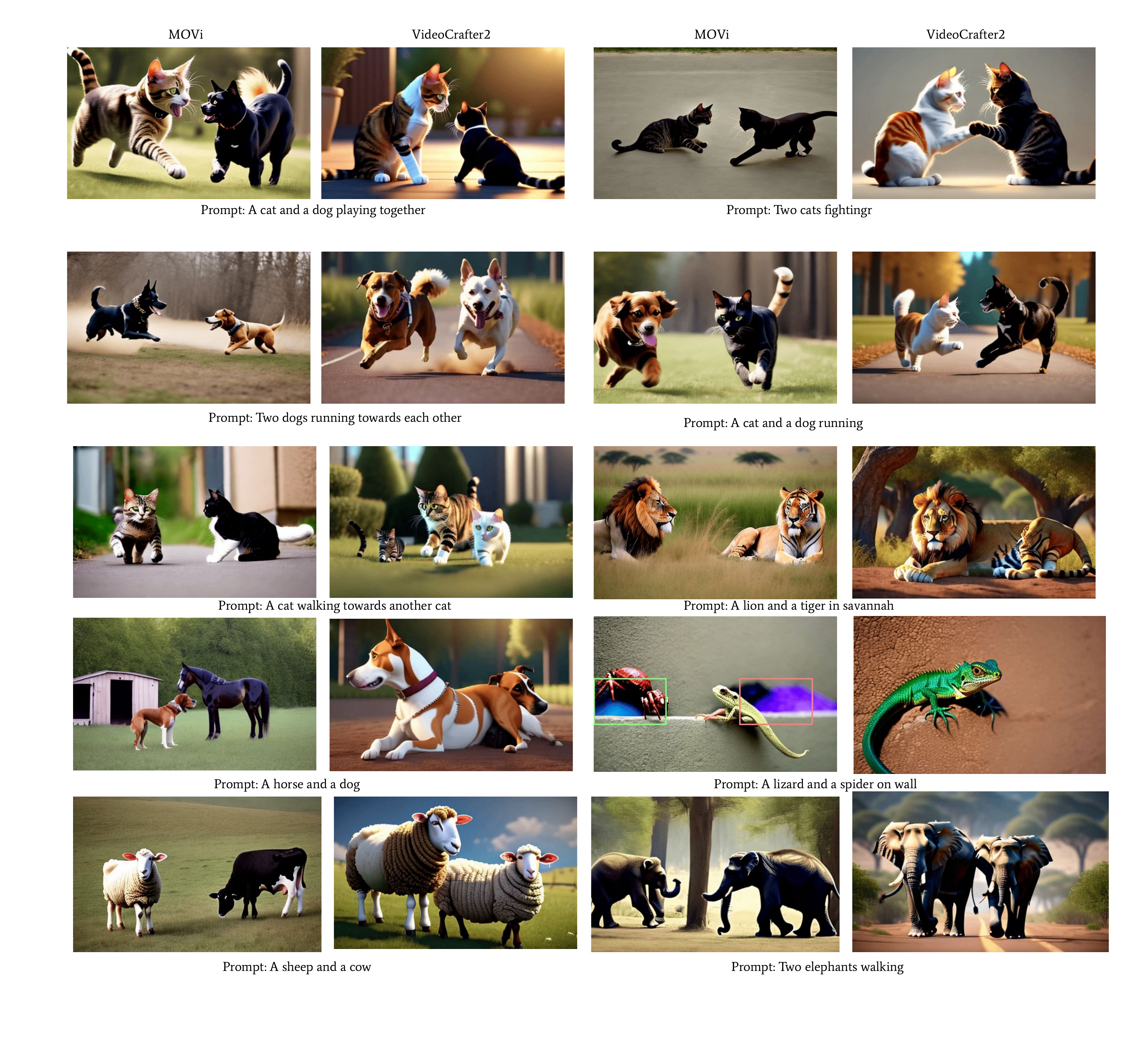}
\caption{Qualitative Comparison with baseline model. The moving version is attached with the supplementary materials as .pptx file.
}

\label{fig:quali}
\end{figure*}

\begin{figure*}[htbp]
\centering
\includegraphics[width=1\linewidth]
{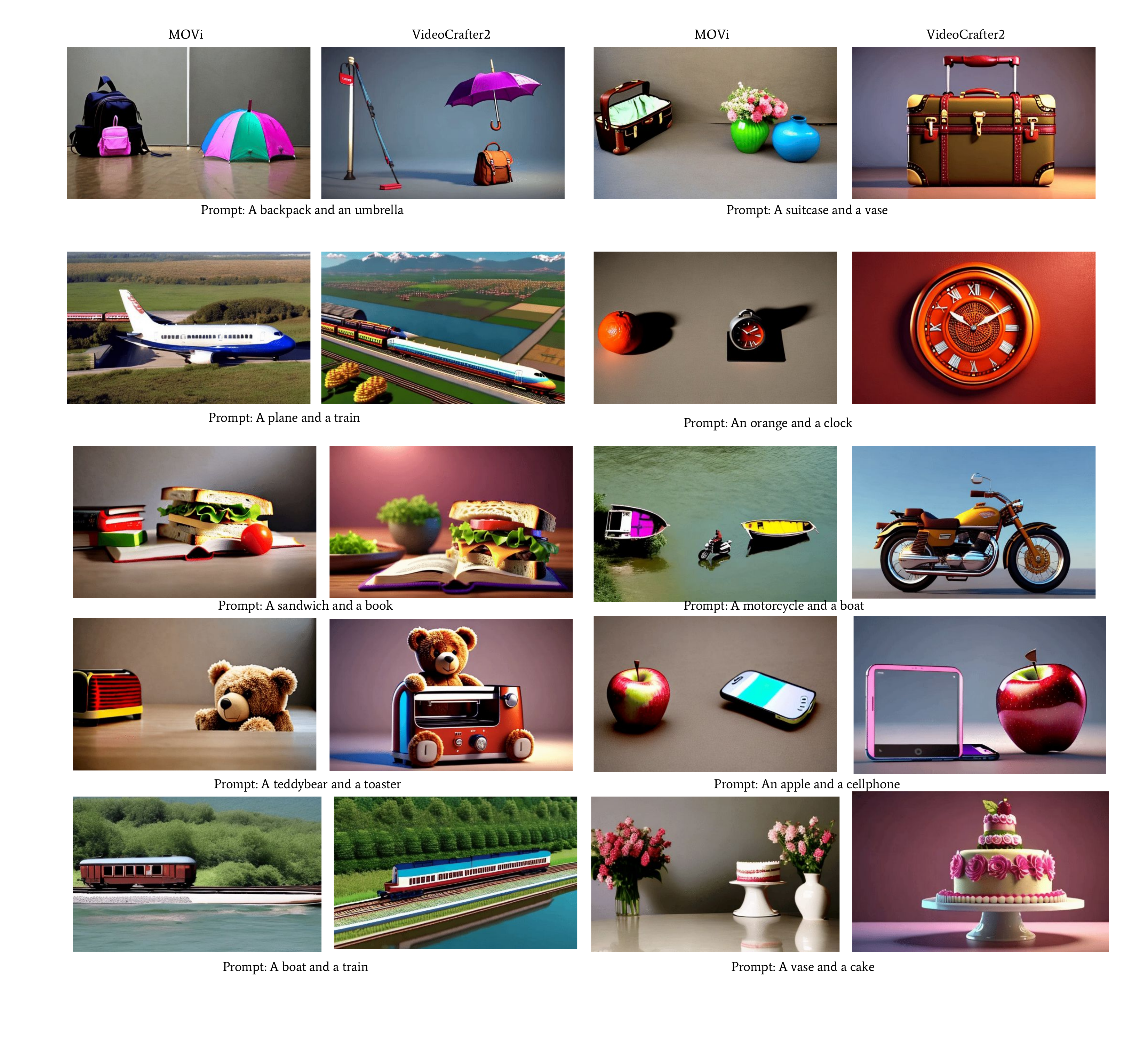}
\caption{Qualitative Comparison with baseline model. The moving version is attached with the supplementary materials as .pptx file.
}

\label{fig:quali2}
\end{figure*}

We provide more qualitative results in~\cref{fig:quali} and ~\cref{fig:quali2}. MOVi is able to better generate multiple distinct objects and more accurately capture their interactions compared to the baseline.

\subsection{Examples of Trajectories from Llama-3.1} 

\textit{Prompt: Think of a trajectory for a 16-frame video, given the prompt. Give me (x,y) coordinates of the objects over time. You should provide co-ordinates based on the number of objects. Format should be: $<$Object1$>$: (frame, x, y) … $<$Object 2$>$: (frame, x , y) The prompt is: $<$prompt$>$}

\noindent\textbf{1) Prompt:} A cat and a dog running.

\noindent\textbf{Response:} Here's a possible trajectory for the cat and dog:
Cat:
(1, 10, 20) (2, 15, 22) (3, 20, 25) (4, 25, 28) (5, 30, 30) (6, 35, 32) (7, 40, 35) (8, 45, 38) (9, 50, 40) (10, 55, 42) (11, 60, 45) (12, 65, 48) (13, 70, 50) (14, 75, 52) (15, 80, 55) (16, 85, 58)
Dog:
(1, 50, 10) (2, 45, 12) (3, 40, 15) (4, 35, 18) (5, 30, 20) (6, 25, 22) (7, 20, 25) (8, 15, 28) (9, 10, 30) (10, 5, 32) (11, 0, 35) (12, -5, 38) (13, -10, 40) (14, -15, 42) (15, -20, 45) (16, -25, 48)
In this trajectory, the cat starts at the top left (10, 20) and moves diagonally down to the right, while the dog starts at the middle right (50, 10) and moves diagonally up to the left. The coordinates are just examples and can be adjusted to fit the specific animation or video.

\noindent\textbf{2) Prompt:} A cat and a dog fighting.

\noindent\textbf{Response:} Here's a possible trajectory for the cat and dog:
Cat:
(1, 40, 40) (2, 42, 38) (3, 45, 35) (4, 48, 32) (5, 50, 30) (6, 52, 28) (7, 55, 25) (8, 58, 22) (9, 60, 20) (10, 62, 18) (11, 60, 20) (12, 58, 22) (13, 55, 25) (14, 52, 28) (15, 50, 30) (16, 48, 32)
Dog:
(1, 60, 60) (2, 58, 62) (3, 55, 65) (4, 52, 68) (5, 50, 70) (6, 48, 72) (7, 45, 75) (8, 42, 78) (9, 40, 80) (10, 38, 82) (11, 40, 80) (12, 42, 78) (13, 45, 75) (14, 48, 72) (15, 50, 70) (16, 52, 68)
In this trajectory, the cat starts at the middle left (40, 40) and moves in a circular motion, as if swiping at the dog. The dog starts at the middle right (60, 60) and also moves in a circular motion, as if swiping back at the cat. The coordinates are just examples and can be adjusted to fit the specific animation or video.

\noindent\textbf{3) Prompt:} A man and a woman skydiving.

\noindent\textbf{Response:} Here's a possible trajectory for the man and woman:
Man:
(1, 50, 100) (2, 52, 95) (3, 55, 90) (4, 58, 85) (5, 60, 80) (6, 62, 75) (7, 65, 70) (8, 68, 65) (9, 70, 60) (10, 72, 55) (11, 75, 50) (12, 78, 45) (13, 80, 40) (14, 82, 35) (15, 85, 30) (16, 88, 25)
Woman:
(1, 70, 100) (2, 68, 95) (3, 65, 90) (4, 62, 85) (5, 60, 80) (6, 58, 75) (7, 55, 70) (8, 52, 65) (9, 50, 60) (10, 48, 55) (11, 45, 50) (12, 42, 45) (13, 40, 40) (14, 38, 35) (15, 35, 30) (16, 32, 25)
In this trajectory, the man and woman start at the top of the frame (100) and move downwards, as if skydiving. The man moves slightly to the right, while the woman moves slightly to the left, as if they are falling together but spreading out horizontally. The coordinates are just examples and can be adjusted to fit the specific animation or video.

\noindent\textbf{4) Prompt:} A frog jumping and rat sitting.

\noindent\textbf{Response:} Here's a possible trajectory for the frog and rat:
Frog:
(1, 40, 80) (2, 42, 75) (3, 45, 70) (4, 50, 65) (5, 55, 60) (6, 60, 55) (7, 65, 50) (8, 70, 45) (9, 75, 40) (10, 80, 35) (11, 85, 30) (12, 80, 35) (13, 75, 40) (14, 70, 45) (15, 65, 50) (16, 60, 55)
Rat:
(1, 80, 20) (2, 80, 20) (3, 80, 20) (4, 80, 20) (5, 80, 20) (6, 80, 20) (7, 80, 20) (8, 80, 20) (9, 80, 20) (10, 80, 20) (11, 80, 20) (12, 80, 20) (13, 80, 20) (14, 80, 20) (15, 80, 20) (16, 80, 20)
In this trajectory, the frog starts at the middle left (40, 80) and jumps in an arc, landing at the middle right (60, 55). The rat remains stationary at the bottom right (80, 20) throughout the 16 frames. The coordinates are just examples and can be adjusted to fit the specific animation or video.

\noindent\textbf{5) Prompt:} Three dogs playing in the field.

\noindent\textbf{Response:} Here's a possible trajectory for the three dogs:
Dog 1:
(1, 20, 40) (2, 25, 35) (3, 30, 30) (4, 35, 25) (5, 40, 20) (6, 45, 15) (7, 50, 10) (8, 55, 5) (9, 50, 10) (10, 45, 15) (11, 40, 20) (12, 35, 25) (13, 30, 30) (14, 25, 35) (15, 20, 40) (16, 15, 45)
Dog 2:
(1, 60, 20) (2, 55, 25) (3, 50, 30) (4, 45, 35) (5, 40, 40) (6, 35, 45) (7, 30, 50) (8, 25, 55) (9, 30, 50) (10, 35, 45) (11, 40, 40) (12, 45, 35) (13, 50, 30) (14, 55, 25) (15, 60, 20) (16, 65, 15)
Dog 3:
(1, 40, 80) (2, 45, 75) (3, 50, 70) (4, 55, 65) (5, 60, 60) (6, 65, 55) (7, 70, 50) (8, 75, 45) (9, 70, 50) (10, 65, 55) (11, 60, 60) (12, 55, 65) (13, 50, 70) (14, 45, 75) (15, 40, 80) (16, 35, 85)
In this trajectory, the three dogs move around the field in different paths, as if playing together. Dog 1 moves in a circular motion, Dog 2 moves in a zig-zag pattern, and Dog 3 moves in a figure-eight pattern. The coordinates are just examples and can be adjusted to fit the specific animation or video.

\noindent\textbf{6) Prompt:}  Three dogs playing and three cats watching.

\noindent\textbf{Response:} Here's a possible trajectory for the 3 dogs and 3 cats:
Dog 1:
(1, 20, 40) (2, 25, 35) (3, 30, 30) (4, 35, 25) (5, 40, 20) (6, 45, 15) (7, 50, 10) (8, 55, 5) (9, 50, 10) (10, 45, 15) (11, 40, 20) (12, 35, 25) (13, 30, 30) (14, 25, 35) (15, 20, 40) (16, 15, 45)
Dog 2:
(1, 60, 20) (2, 55, 25) (3, 50, 30) (4, 45, 35) (5, 40, 40) (6, 35, 45) (7, 30, 50) (8, 25, 55) (9, 30, 50) (10, 35, 45) (11, 40, 40) (12, 45, 35) (13, 50, 30) (14, 55, 25) (15, 60, 20) (16, 65, 15)
Dog 3:
(1, 40, 80) (2, 45, 75) (3, 50, 70) (4, 55, 65) (5, 60, 60) (6, 65, 55) (7, 70, 50) (8, 75, 45) (9, 70, 50) (10, 65, 55) (11, 60, 60) (12, 55, 65) (13, 50, 70) (14, 45, 75) (15, 40, 80) (16, 35, 85)
Cat 1:
(1, 80, 40) (2, 80, 40) (3, 80, 40) (4, 80, 40) (5, 80, 40) (6, 80, 40) (7, 80, 40) (8, 80, 40) (9, 80, 40) (10, 80, 40) (11, 80, 40) (12, 80, 40) (13, 80, 40) (14, 80, 40) (15, 80, 40) (16, 80, 40)
Cat 2:
(1, 90, 60) (2, 90, 60) (3, 90, 60) (4, 90, 60) (5, 90, 60) (6, 90, 60) (7, 90, 60) (8, 90, 60) (9, 90, 60) (10, 90, 60) (11, 90, 60) (12, 90, 60) (13, 90, 60) (14, 90, 60) (15, 90, 60) (16, 90, 60)
Cat 3:
(1, 70, 90) (2, 70, 90) (3, 70, 90) (4, 70, 90) (5, 70, 90) (6, 70, 90) (7, 70, 90) (8, 70, 90) (9, 70, 90) (10, 70, 90) (11, 70, 90) (12, 70, 90) (13, 70, 90) (14, 70, 90) (15, 70, 90) (16, 70, 90)
In this trajectory, the three dogs move around the field playing together, while the three cats remain stationary, watching from different parts of the field. The coordinates are just examples and can be adjusted to fit the specific animation or video.

\noindent\textbf{7) Prompt:}  a frog is running and rat is running.

\noindent\textbf{Response:} You want to see a trajectory for a frog and a rat running together! Here it is:
Frog:
(1, 20, 100) (2, 25, 95) (3, 30, 90) (4, 35, 85) (5, 40, 80) (6, 45, 75) (7, 50, 70) (8, 55, 65) (9, 60, 60) (10, 65, 55) (11, 70, 50) (12, 75, 45) (13, 80, 40) (14, 85, 35) (15, 90, 30) (16, 95, 25)
Rat:
(1, 80, 50) (2, 75, 55) (3, 70, 60) (4, 65, 65) (5, 60, 70) (6, 55, 75) (7, 50, 80) (8, 45, 85) (9, 40, 90) (10, 35, 95) (11, 30, 100) (12, 25, 105) (13, 20, 110) (14, 15, 115) (15, 10, 120) (16, 5, 125)
In this trajectory, the frog is running from left to right, while the rat is running from right to left. Both are moving at different speeds and angles, creating a fun and dynamic scene!

\noindent\textbf{8) Prompt:} Two men walking towards each other.

\noindent\textbf{Response:} Here's a possible trajectory for the two men walking towards each other:
Man 1:
(1, 20, 100) (2, 22, 100) (3, 24, 100) (4, 26, 100) (5, 28, 100) (6, 30, 100) (7, 32, 100) (8, 34, 100) (9, 36, 100) (10, 38, 100) (11, 40, 100) (12, 42, 100) (13, 44, 100) (14, 46, 100) (15, 48, 100) (16, 50, 100) (17, 52, 100) (18, 54, 100) (19, 56, 100) (20, 58, 100) (21, 60, 100) (22, 62, 100) (23, 64, 100) (24, 66, 100) (25, 68, 100) (26, 70, 100) (27, 72, 100) (28, 74, 100) (29, 76, 100) (30, 78, 100) (31, 80, 100) (32, 82, 100)
Man 2:
(1, 180, 100) (2, 178, 100) (3, 176, 100) (4, 174, 100) (5, 172, 100) (6, 170, 100) (7, 168, 100) (8, 166, 100) (9, 164, 100) (10, 162, 100) (11, 160, 100) (12, 158, 100) (13, 156, 100) (14, 154, 100) (15, 152, 100) (16, 150, 100) (17, 148, 100) (18, 146, 100) (19, 144, 100) (20, 142, 100) (21, 140, 100) (22, 138, 100) (23, 136, 100) (24, 134, 100) (25, 132, 100) (26, 130, 100) (27, 128, 100) (28, 126, 100) (29, 124, 100) (30, 122, 100) (31, 120, 100) (32, 118, 100)
In this trajectory, the two men are walking towards each other at a constant speed of 2 pixels/frame. Man 1 is walking from left to right, while Man 2 is walking from right to left. They will meet in the middle of the frame at around frame 16.

\section{Prompt for GPT-4 trajectory evaluation}

\noindent \textbf{Prompt:} Here's a possible trajectory for a 16-frame video of $<$prompt:$>$ a man and a woman skydiving together:

$<$Trajectories:$>$
Man: (1, 100, 100) (2, 120, 120) (3, 140, 140) (4, 160, 160) (5, 180, 180) (6, 200, 200) (7, 220, 220) (8, 240, 240) (9, 260, 260) (10, 280, 280) (11, 300, 300) (12, 320, 320) (13, 340, 340) (14, 360, 360) (15, 380, 380) (16, 400, 400)

Woman: (1, 80, 80) (2, 100, 100) (3, 120, 120) (4, 140, 140) (5, 160, 160) (6, 180, 180) (7, 200, 200) (8, 220, 220) (9, 240, 240) (10, 260, 260) (11, 280, 280) (12, 300, 300) (13, 320, 320) (14, 340, 340) (15, 360, 360) (16, 380, 380)

Rate this trajectory, consider realisticness, physics, smoothness and consistency. Give a combined score only.

\section{Prompt Generated by GPT-4} Vbench multi-object prompts do not have any action specified. We generate multi-object prompts with action using GPT-4 for dynamic degree analysis of the generated videos. Below are the list of prompts:\\
A bird and a cat sitting.\\
A cat and a dog running.\\
A dog and a horse walking.\\
A horse and a sheep grazing.\\
A sheep and a cow standing.\\
A cow and an elephant drinking.\\
An elephant and a bear walking.\\
A bear and a zebra running.\\
A zebra and a giraffe eating.\\
A giraffe and a bird looking.\\
An airplane and a train moving.\\
A train and a boat traveling.\\
A boat and a bicycle floating.\\
A bicycle and a car riding.\\
A car and a motorcycle speeding.\\
A motorcycle and a bus stopping.\\
A bus and a truck driving.\\
A truck and a person walking.\\
A person and a bird climbing.\\
A bird and a dog playing.\\

\section{Discussions}
\subsection{Limitations}
\begin{figure*}[htbp]
\centering
\includegraphics[width=1\linewidth]
{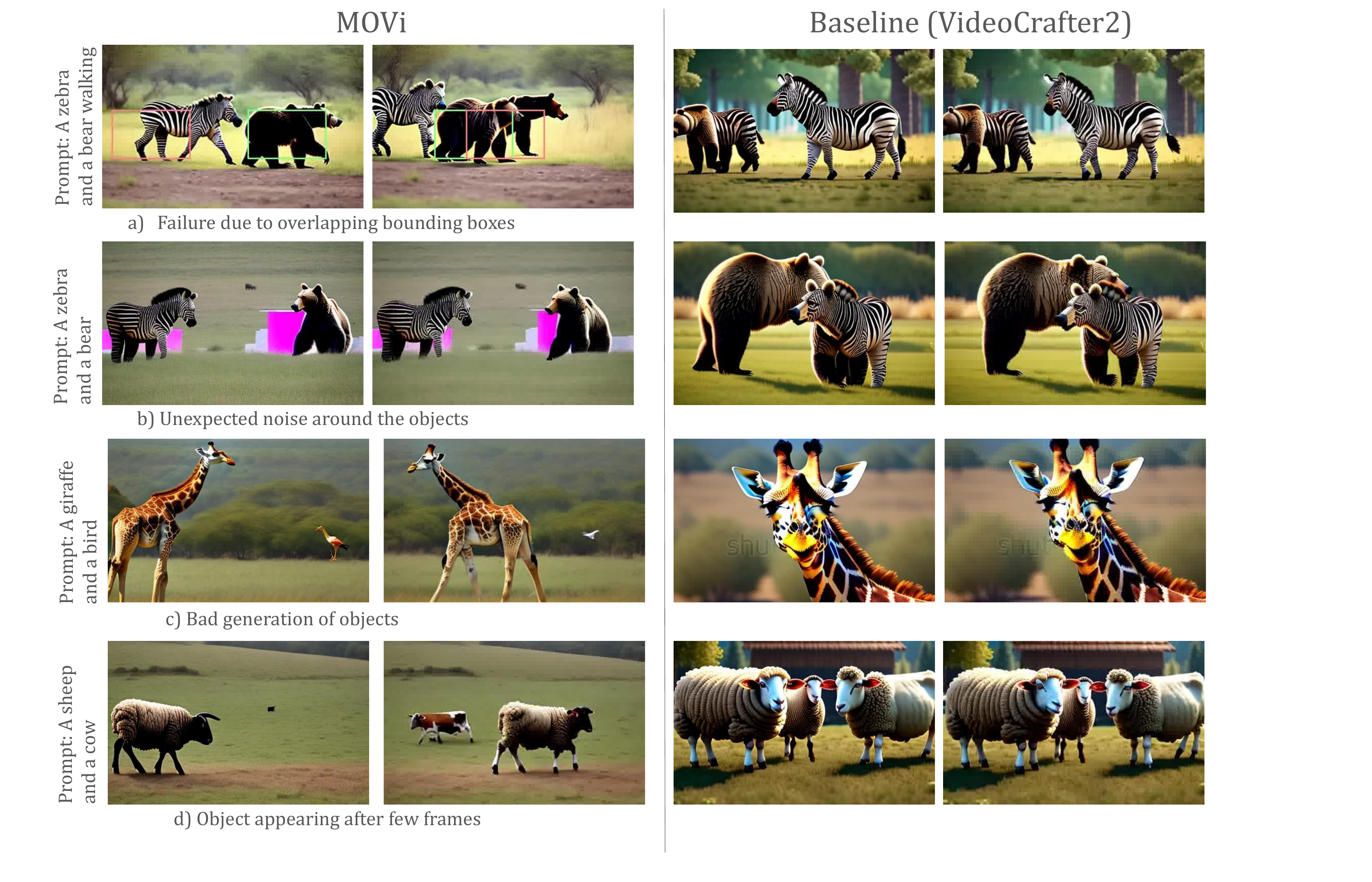}
\caption{Examples of failure cases are shown: (a) Overlapping bounding boxes may lead to anomalies. (b) Generated outputs might include additional noise around the bounding boxes. (c) Individual object generation failures could occur, potentially reflecting limitations of the base model. (d) Some objects may appear later in the video, which, while not inherently problematic, could impact frame-by-frame object detection performance. For reference, outputs from the base model are also provided.
}
\label{Fig:fail}
\end{figure*}

Our approach relies on the prior world knowledge embedded in the video diffusion models and LLMs, which helps align the generated trajectories with their corresponding objects. However, since we do not explicitly label each trajectory with its intended object, the model's performance is contingent on this implicit alignment. While this typically works well in practice, it may fail in ambiguous cases where the model's priors of a particular object movement is insufficient. Additionally, the overall quality of the generation is inherently tied to the capabilities of the base models. If the initial generation is suboptimal, it can propagate errors throughout the process, leading to poor final outputs. Some examples of limitation are illustrated in Figure \ref{Fig:fail}.

\subsection{Negative Societal Impact}
The multi-object, training-free video generation models we present have strong potential for creative applications, offering quick video generation based on input text. While not all outputs will precisely match the intent, they can still significantly reduce content creation time. However, this ease of use also brings ethical concerns. The models could be misused to generate misleading or harmful content, especially as video quality improves. Ensuring responsible use is essential as this technology develops further.

    

    

\end{document}